\documentclass[letterpaper]{article} % DO NOT CHANGE THIS
\usepackage{aaai2026}  % DO NOT CHANGE THIS
\usepackage{times}  % DO NOT CHANGE THIS
\usepackage{helvet}  % DO NOT CHANGE THIS
\usepackage{courier}  % DO NOT CHANGE THIS
\usepackage[hyphens]{url}  % DO NOT CHANGE THIS
\usepackage{graphicx} % DO NOT CHANGE THIS
\usepackage{natbib}  % DO NOT CHANGE THIS AND DO NOT ADD ANY OPTIONS TO IT
\usepackage{caption} % DO NOT CHANGE THIS AND DO NOT ADD ANY OPTIONS TO IT
\usepackage{algorithm}
\usepackage{algorithmic}

\usepackage{newfloat}
\usepackage{listings}
\DeclareCaptionStyle{ruled}{labelfont=normalfont,labelsep=colon,strut=off} % DO NOT CHANGE THIS
\floatstyle{ruled}
\newfloat{listing}{tb}{lst}{}
\floatname{listing}{Listing}
\usepackage{booktabs}
\usepackage{amsmath}
\usepackage{amsfonts}

\title{Enhancing Conversational Recommender Systems with Tree-Structured Knowledge and Pretrained Language Models}
\author{
    Yongwen Ren\textsuperscript{\rm 1,\rm 2},
    Chao Wang\textsuperscript{\rm 3}\thanks{Corresponding author.},
    Peng Du\textsuperscript{\rm 4},
    Chuan Qin\textsuperscript{\rm 5,\rm 6},
    Dazhong Shen\textsuperscript{\rm 7},
    Hui Xiong\textsuperscript{\rm 8,\rm 9}\footnotemark[1]
}
\affiliations{
    \textsuperscript{\rm 1}School of Computer Science and Technology, University of Science and Technology of China\\
    \textsuperscript{\rm 2}iFLYTEK AI Research, iFLYTEK Co.,Ltd\\
    \textsuperscript{\rm 3}School of Artificial Intelligence and Data Science, University of Science and Technology of China\\
    \textsuperscript{\rm 4}School of Software and Microelectronics, Peking University\\
    \textsuperscript{\rm 5}Computer Network Information Center, Chinese Academy of Sciences\\
    \textsuperscript{\rm 6}University of Chinese Academy of Sciences\\
    \textsuperscript{\rm 7}College of Computer Science and Technology, Nanjing University of Aeronautics and Astronautics\\
    \textsuperscript{\rm 8}Thrust of Artificial Intelligence, The Hong Kong University of Science and Technology (Guangzhou)\\
    \textsuperscript{\rm 9}Department of Computer Science and Engineering, The Hong Kong University of Science and Technology\\
    yovren@mail.ustc.edu.cn, 
    wangchaoai@ustc.edu.cn, 
    pdu@pku.edu.cn,\\
    chuanqin0426@gmail.com, 
    shendazhong@nuaa.edu.cn, 
    xionghui@ust.hk
}

\usepackage{bibentry}
\begin{document}

\maketitle

\begin{abstract}

    Recent advances in pretrained language models (PLMs) have significantly improved conversational recommender systems (CRS), enabling more fluent and context-aware interactions.
    To further enhance accuracy and mitigate hallucination, many methods integrate PLMs with knowledge graphs (KGs), but face key challenges: failing to fully exploit PLM reasoning over graph relationships, indiscriminately incorporating retrieved knowledge without context filtering, and neglecting collaborative preferences in multi-turn dialogues.
    To this end, we propose PCRS-TKA, a prompt-based framework employing retrieval-augmented generation to integrate PLMs with KGs.
    PCRS-TKA constructs dialogue-specific knowledge trees from KGs and serializes them into texts, enabling structure-aware reasoning while capturing rich entity semantics.
    Our approach selectively filters context-relevant knowledge and explicitly models collaborative preferences using specialized supervision signals.
    A semantic alignment module harmonizes heterogeneous inputs, reducing noise and enhancing accuracy.
    Extensive experiments demonstrate that PCRS-TKA consistently outperforms all baselines in both recommendation and conversational quality.

\end{abstract}

\begin{links}
    \link{Code}{https://github.com/YovRen/PCRS-TKA}
\end{links}

\section{Introduction}
Recommendation systems play a crucial role in intelligent assistants by helping users efficiently discover relevant items. However, traditional systems lack interactive dialogue abilities, limiting flexibility and explainability~\cite{chenSurveyDialogueSystems2017,ji2025comprehensive}. Conversational recommender systems (CRS) address this issue by supporting personalized and natural interactions. With recent advances in pretrained language models (PLMs), CRS have achieved notable improvements in conversational fluency and context understanding~\cite{wuEmpoweringNewsRecommendation2021,qin2025scihorizon}.

However, PLMs still suffer from hallucinations, generating inaccurate or irrelevant information, which damages recommendation reliability~\cite{HallucinationDetectionRobustly,zhangSirensSongAI2023}. To mitigate this issue, knowledge graphs (KGs) can be integrated with PLMs to provide factual external knowledge, thereby improving both accuracy and robustness of conversational recommendations~\cite{wangUnifiedConversationalRecommender2022,wang2018confidence,wang2025unleashing}.

In the literature, early CRS primarily relied on structured conversations centered around item attributes such as genre or price~\cite{gaoAdvancesChallengesConversational2021}. Later, KG-based CRS~\cite{wangKGATKnowledgeGraph2019,petroniLanguageModelsKnowledge2019,bouraouiInducingRelationalKnowledge2020} integrated external knowledge resources and developed alignment techniques to ensure semantic consistency, e.g., {KBRD}~\cite{chenKnowledgeBasedRecommenderDialog2019} leveraging relational graph convolutional networks (RGCNs)~\cite{schlichtkrullModelingRelationalData2017}, {KGSF}~\cite{zhouImprovingConversationalRecommender2020} applying mutual information maximization for word-entity alignment.
More recent PLM-based approaches, including {BARCOR}~\cite{wangBARCORUnifiedFramework2022}, {UniCRS}~\cite{wangUnifiedConversationalRecommender2022}, {KERL}~\cite{Qiu_2025}, and {DCRS}~\cite{dao_broadening_2024}, combine prompt learning with KG integration to mitigate hallucination and boost domain knowledge. However, these existing methods fail to fully exploit the rich semantic information embedded in KGs, limiting their ability to capture complex relational patterns and contextual dependencies.

Despite progress, several challenges persist in integrating PLMs and KGs for conversational recommendation. First, many existing approaches depend on GCN-based encoders (e.g., RGCN)~\cite{zhu2024graph}, which struggle to leverage PLMs' reasoning capabilities over complex graph structures and cannot fully capture rich KG semantics~\cite{shen2021topic,wang2024unleashing}.
Second, current approaches typically integrate all retrieved KG information without filtering out information irrelevant to the current dialogue context. This inevitably introduces noise and compromises the model's performance.
Third, dialogue text is often treated merely as textual input data, neglecting valuable latent user collaborative preference information embedded in conversations. This oversight can result in suboptimal personalized recommendations that fail to align with users' true preferences~\cite{wang2021personalized}.

To address these issues, we propose PCRS-TKA, a prompt-based framework that integrates PLMs with KGs via retrieval-augmented generation (RAG) style strategy.
First, to augment the static, global embeddings from RGCN and fully exploit PLM reasoning capabilities, PCRS-TKA constructs dynamic, context-aware knowledge trees in a RAG-style. These trees are selectively built from the KG, serialized into text, and fed into the PLM to enable direct, fine-grained reasoning on dialogue-specific structured information.
Second, our approach filters context-relevant semantics through this dialogue-specific knowledge tree construction, ensuring only pertinent information is utilized.
Third, we explicitly model collaborative preferences from multi-turn dialogues using specialized supervision signals, capturing latent user preference patterns that are often overlooked in existing methods.
Additionally, we employ an RGCN to encode the global semantics of the entire KG, and these multiple sources of knowledge are jointly embedded into prompt representations, enabling the PLM to perform end-to-end graph reasoning in a unified manner.
Complemented by a semantic alignment module that harmonizes heterogeneous inputs from dialogues and KGs, PCRS-TKA effectively reduces noise and enhances recommendation accuracy. Extensive experiments demonstrate its superiority in both recommendation and conversational quality.

\section{Related Work}
\subsection{Conversational Recommendation Systems}
Early CRS methods focused on structured conversations that gathered user preferences through item attributes like genre or price, using predefined templates and algorithms such as multi-armed bandits or reinforcement learning~\cite{wang2021variable}. However, these methods lacked flexibility and natural language generation capabilities.
KG-based CRS methods were later developed to improve this. {KBRD}~\cite{chenKnowledgeBasedRecommenderDialog2019} introduced KGs and RGCNs to model relationships between items and users. {KGSF}~\cite{zhouImprovingConversationalRecommender2020} enhanced this by incorporating word-level KGs and MIM to align word and entity representations, resulting in more coherent responses. {RevCore}~\cite{luRevCoreReviewAugmentedConversational2021} enriched dialogue generation with unstructured review data, while {C2CRS}~\cite{zhouC2CRSCoarseFineContrastive2023} used multi-granularity contrastive learning for multimodal data alignment~\cite{zong2024mova}. Despite these advancements, KG-based methods often treated recommendation and dialogue modules separately, limiting the full use of dialogue content.
With the rise of PLMs, prompt learning was introduced to improve conversational capabilities. For example, {BARCOR}~\cite{wangBARCORUnifiedFramework2022} uses BART~\cite{lewisBARTDenoisingSequenceSequence2020} for better response generation, {UniCRS}~\cite{wangUnifiedConversationalRecommender2022} integrates recommendation and dialogue generation via prompt learning, and {DCRS}~\cite{dao_broadening_2024} employs a knowledge-aware retriever for improved capabilities.

\subsection{Unifying PLMs and KGs in Recommendations}
Recommendation systems often require domain-specific knowledge that PLMs alone cannot provide. KGs supply structured knowledge to fill this gap. Many studies focus on aligning the semantic spaces of PLMs and KGs by modifying Transformer architectures, often incorporating cross-attention mechanisms to jointly model dialogue text and KG information. For instance, {KGSF}~\cite{zhouImprovingConversationalRecommender2020} uses mutual information maximization for embedding alignment, while {C2CRS}~\cite{zhouC2CRSCoarseFineContrastive2023} applies contrastive learning at both sentence and word levels to improve semantic consistency.
In prompt learning frameworks, the PLM architecture generally remains fixed, and KG information is integrated by concatenating implicit graph embeddings derived from graph neural networks (GNNs) into input prompts, as seen in {UniCRS} and {DCRS}. Although this enables KG data incorporation, it does not fully exploit the reasoning abilities of PLMs over KG relations, nor does it provide dynamic knowledge filtering tailored to the dialogue context.

\begin{figure*}[ht]
    \centering
    \includegraphics[width=\textwidth]{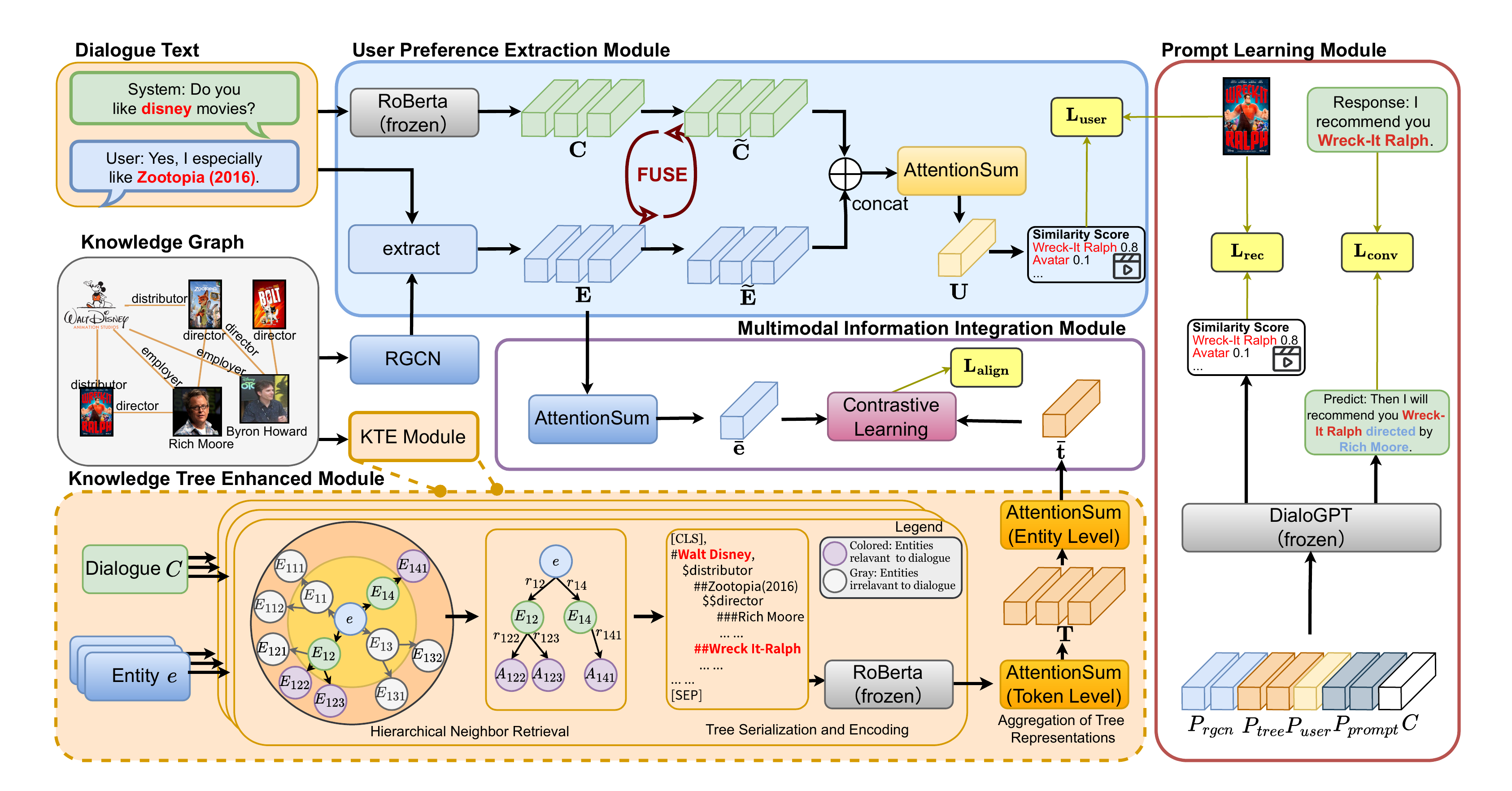}
    \caption{The network architecture of the PCRS-TKA framework.}
    \label{fig:framework}
\end{figure*}

\section{Methodology}

\subsection{Task Formulation}
The goal of CRS is to recommend relevant items while maintaining a natural, interactive dialogue with the user. At each turn,  the system analyzes the dialogue history, infers the user's preferences, and generates a response that includes recommended items within the natural language utterance.
Formally, let \(u\) represent a user in the user set \(\mathcal{U}\), \(i\) an item in the item set \(I\), and \(w\) a word in the vocabulary \(\mathcal{V}\). A multi-turn conversation \(C\) is defined as a sequence of utterances: $C = \{s_t\}_{t=1}^{n}$, where \(s_t\) denotes the \(t\)-th sentence. Additionally, an external KG \(G\) is represented as a set of triples: $G = \{(e_1, r, e_2)\}$, where \(e_1, e_2 \in \mathcal{E}\) and \(r \in \mathcal{R}\). Here, we assume \(I \subset \mathcal{E}\), indicating that all candidate items are included in \(\mathcal{E}\). Given an \(n\)-turn conversation \(C\), the task of CRS is to generate a response sentence \(s_{n+1}\) and select a set of recommended items \(I_{n+1}\) from the item set~\(I\).

\subsection{Overview of the Approach}
As illustrated in Figure~\ref{fig:framework}, we propose \textbf{PCRS-TKA}, a conversational recommendation framework that unifies PLMs and KGs via knowledge-enhanced prompt learning. The key idea is to augment the PLM with two distinct forms of knowledge: a static, global understanding of the entire KG, and a dynamic, dialogue-specific subgraph for fine-grained reasoning. We also introduce a collaborative signal to guide the PLM's optimization. This is realized through four core components:
1) \textbf{User Preference Extraction Module.} We first employ RoBERTa~\cite{liuRoBERTaRobustlyOptimized2019} , a bidirectional PLM, and RGCN~\cite{schlichtkrullModelingRelationalData2017} to encode dialogue text \( C \) and mentioned KG entities \( E \) into embeddings $\mathbf{C}$ and $\mathbf{E}$, user preference features \( \mathbf{U} \) are then extracted explicitly and supervised by  an auxiliary recommendation task to ensure that \( P_{\text{user}} \) effectively learns and incorporates this information.
2) \textbf{Knowledge Tree Enhanced Module.} For each entity in \( E \), we retrieve a multi-hop knowledge tree from \( G \) using a RAG-style strategy. These trees are then hierarchically aggregated based on structural and contextual relevance, forming a serialized textual sequence \( T \), which is encoded into \( P_{\text{tree}} \).
3) \textbf{Multimodal Information Integration Module} This module aligns $\mathbf{T}$ with $\mathbf{E}$ through pair-wise contrastive learning, effectively integrating information across modules.
4) \textbf{Prompt Learning Module.} The concatenated prompt sequence  \((P_{\text{prompt}}, P_{\text{rgcn}}, P_{\text{tree}}, P_{\text{user}})\), including learnable soft prompts \( P_{\text{prompt}} \), is prepended to the dialogue context and fed into DialoGPT.

\subsection{User Preference Extraction Module}\label{User}
This module learns user preference representations $\mathbf{U}$ from multi-turn dialogue $C$ and injects them into the PLM as an auxiliary supervision signal for prompt learning.

1) \textbf{Feature Encoder.} For natural language understanding, we use RoBERTa to encode the dialogue text $C$ into embeddings $\mathbf{C} $ as follows:
{\small\begin{align}
    \mathbf{C} = \mathrm{RoBERTa}(C).
\end{align}}%
To provides structured information about the entities mentioned in the conversation, we incorporate a knowledge graph \( G \) and adopt RGCN to encode it into embeddings \( \mathbf{G} \), the embeddings  \( \mathbf{E} \) corresponding to $E$ are retrieved from matrix $\mathbf{G}$ as below:
{\small\begin{align}
     & \mathbf{G} = \mathrm{RGCN}(G),                 \\
     & \mathbf{E} = \mathrm{Retrieve}(\mathbf{G}, E).
\end{align}}%

2) \textbf{User Collaborative Information Extraction.} Since $\mathbf{C} $ and $\mathbf{E} $ originate from different sources, a semantic gap exists between them. To bridge this gap, we employ a cross-interaction mechanism~\cite{wangUnifiedConversationalRecommender2022} to align their semantic spaces via linear transformations and bilinear interaction:
{\small\begin{align}
     & \mathbf{C} = \mathbf{C} W_C, \mathbf{E} = \mathbf{E} W_E,\mathbf{A} = \mathbf{C} W \mathbf{E}^{\top},                            \\
     & \mathbf{\widetilde{C}} = \mathbf{C} + \mathbf{E} \mathbf{A}, \mathbf{\widetilde{E}} = \mathbf{E} + \mathbf{C} \mathbf{A}^{\top},
\end{align}}%
This interaction fuses contextual semantics from the dialogue and relational knowledge from the KG, generating  enriched representations \( \mathbf{\widetilde{C}}\) and \( \mathbf{\widetilde{E}} \).
Furthermore,  we concatenate them and apply a self-attention mechanism to capture dependencies across the entire input sequence. We define the resulting attention aggregation as a function $\text{ASum}(\cdot)$, and compute the user preference embedding $\mathbf{U}$ as follows:
{\small\begin{align}
    \text{ASum}(\mathbf{X}) & = \sum_{i=1}^{n_X} \sum_{j=1}^{n_X} \text{softmax} \left( \frac{\mathbf{X}_i W_Q \cdot (\mathbf{X}_j W_K)^\top}{\sqrt{d}} \right) \cdot \mathbf{X}_j W_V, \\
    \mathbf{X}              & = \text{concat}(\mathbf{\widetilde{C}}, \mathbf{\widetilde{E}}), \quad \mathbf{U} = \text{ASum}(\mathbf{X})
\end{align}}%
\noindent where $W_{Q}, W_{K}, W_{V}$ are learnable parameters.

3) \textbf{Supervision Recommendation Task.} To introduce collaborative preference supervision, we define an auxiliary task. Instead of implicitly fetching neighbor features~\cite{xiezhong2024,liuyuan2025}, this task learns to explicitly encode a collaborative signal by computing relevance scores between the user embedding $\mathbf{U}$ and candidate item embeddings $\mathbf{I}$, which are retrieved from the graph entity embeddings $\mathbf{G}$. We then project $\mathbf{I}$ and compute the scores:
{\small\begin{align}
    \mathbf{I} = \mathrm{Retrieve}(\mathbf{G}, I) ,\quad\mathbf{I} = \mathbf{I}W_I,\quad
    R = \operatorname{softmax}(\mathbf{U} \mathbf{I}^{\top}),
\end{align}}%
\noindent where $W_I$ is a learnable projection matrix and $ R$ represents the predicted rating scores. Given \(N\) conversations and ground-truth preference labels $\mathbf{Y} $ , we can compute the predicted rating score $ R $ in all conversations, and the cross-entropy loss is defined as:
{\small\begin{align}
    \! L_{user} \! & =\! -\sum_{j=1}^{N} \sum_{i=1}^{n_I} \left[Y_j^i \cdot \log R_j^i + (1 - Y_j^i) \cdot \log (1 - R_j^i)\right].
\end{align}}%

\subsection{Knowledge Tree Enhanced Module}\label{Tree}
To better utilize the structured semantics in KGs, we dynamically construct a dialogue-specific knowledge tree that captures context-relevant information in a hierarchical form, enabling PLMs to conduct structure-aware reasoning.

1) \textbf{Hierarchical Neighbor Retrieval.} Given a dialogue context $C$ and the mentioned entities $E = \{e_1, \dots, e_{n_E}\}$, we construct a personalized knowledge tree $G_{\text{tree}_e}$ for each entity $e \in E$. We first encode the dialogue into a fixed-length vector by applying a pooling operation (e.g., mean pooling) over word embeddings:
{\small\begin{align}
    \mathbf{c} = \operatorname{Pooling}(\mathbf{C}).
\end{align}}%
Starting from the root $\mathcal{N}_e^0 = \{e\}$, each layer $l$ expands by retrieving one-hop neighbors $\mathcal{M}(v)$ for every node $v \in \mathcal{N}_e^l$. Their relevance to the dialogue is measured via cosine similarity and we select the top-$N$ most relevant neighbors to form the next layer until reaching depth $L$:
{\small\begin{align}
    \mathrm{sim}(\mathbf{c}, \mathbf{E}_m) = & \frac{\mathbf{c} \cdot \mathbf{E}_m^\top}{\|\mathbf{c}\| \|\mathbf{E}_m\|},             \\
    \mathcal{N}_v^{l+1} =                    & \operatorname{TopN}\big(\mathcal{M}(v), \mathrm{sim}(\mathbf{c}, \mathbf{E}_m), N\big), \\
    \mathcal{N}_e^{l+1} =                    & \bigcup_{v \in \mathcal{N}_e^l} \mathcal{N}_v^{l+1},                                    \\
    G_{\text{tree}_e} =                      & \left(\bigcup_{l=0}^{L} \mathcal{N}_e^l, \text{corresponding edges}\right).
\end{align}}%

2) \textbf{Tree Serialization and Encoding.} To retain structural information, virtual relation nodes are inserted between parent–child pairs. The tree is serialized into a natural language sequence via depth-first traversal, where entities and relations are formatted as ``\#EntityName'' or ``\$RelationName'' to indicate depth. The serialized sequence $T_e$ is encoded using RoBERTa to produce contextualized token embeddings:
{\small\begin{align}
    \mathbf{T}_e = \mathrm{RoBERTa}(T_e).
\end{align}}%

3) \textbf{Aggregation of Tree Representations.} To obtain a condensed representation for each tree, we apply token-level self-attention followed by weighted summation to get the representation of $T_e$. The entity-level embeddings for all trees are then concatenated as follows:
{\small
\begin{align}
    \mathbf{t}_e & =  \text{ASum}(\mathbf{T}_e),                            \\
    \mathbf{T}   & = [\mathbf{t}_1; \mathbf{t}_2; \dots; \mathbf{t}_{n_E}].
\end{align}}%
A further self-attention layer captures interactions among entities and yields a unified knowledge representation. The final vector $\mathbf{t}_E$ encodes structured, dialogue-specific knowledge for downstream recommendation and generation:
{\small\begin{align}
    \mathbf{t}_E = \text{ASum}(\mathbf{T}).
\end{align}}%

\subsection{Multimodal Information Integration Module}
Our framework incorporates information from both conversations and KGs. The KG provides two complementary forms of embeddings: entity embeddings $\mathbf{E}$ and knowledge tree embeddings $\mathbf{T}$. Aligning them helps ensure that representations of the same entity remain close in the semantic space. Since the entity embedding matrix $\mathbf{E} = [\mathbf{e}_1;\mathbf{e}_2;\dots;\mathbf{e}_{n_E}]$ contains all mentioned entities, we obtain a single entity-level representation using the same attention-based aggregation introduced in Section~\ref{Tree}:
{\small\begin{align}
    \mathbf{e}_E = \text{ASum}(\mathbf{E}).
\end{align}}%

To align  $\mathbf{e}_E$ and $\mathbf{t}_E$, we employ contrastive learning\cite{maNoiseContrastiveEstimation2018}, encouraging positive pairs to be close and negative pairs to be separated. Specifically, given two entity sequences $E_i$ and $E_j$ from a mini-batch, $\mathbf{e}_{E_i}$ and $\mathbf{t}_{E_j}$ form a positive pair if $E_i = E_j$, and a negative pair otherwise. For a batch of $b$ conversations with sequences $[E_1, \dots, E_b]$, the contrastive loss is:
{\small
\begin{align}
    M_{i, j}        & = \mathbb{I}[E_i = E_j],                                                                                     \\
    L_{\text align} & = \sum_{i=1}^{b}\sum_{j=1}^{b} -\log\frac{e^{(\mathbf{e}_{E_i}\cdot \mathbf{t}_{E_j}) / \tau} \cdot M_{i,j}}
    {\sum_{k=1}^b (e^{(\mathbf{e}_{E_i}\cdot \mathbf{t}_{E_k}) / \tau} \cdot (1-M_{i,k}))},
\end{align}
}%
where $\tau$ is a temperature hyperparameter that controls the sharpness of similarity discrimination.

\subsection{Prompt Learning Module}\label{Prompt}
The overall model is divided into four parameter groups: the base PLM $\Theta_{\text{plm}}$, the user preference extraction module $\Theta_{\text{user}}$, the knowledge tree enhanced module $\Theta_{\text{tree}}$, and the task-specific soft prompts $\Theta_{\text{prompt}}$. We adopt DialoGPT~\cite{zhangDialoGPTLargeScaleGenerative2019}, a Transformer-based autoregressive model pretrained on large-scale Reddit dialogues, as the backbone PLM. Its parameters $\Theta_{\text{plm}}$ remain frozen during training, while the other modules are learnable. To reduce hallucinations in recommendation, we follow UniCRS~\cite{wangUnifiedConversationalRecommender2022} and append the generated conversation response to the dialogue context $C$ when constructing the recommendation input. The final inputs for generation and recommendation tasks are:
{\small\begin{align}
    \widetilde{C}_{\text{gen}} & = \text{concat}(P_{\text{rgcn}}, P_{\text{tree}}, P_{\text{user}}, P_{\text{prompt(C)}}, C), \\
    \widetilde{C}_{\text{rec}} & = \text{concat}(P_{\text{rgcn}}, P_{\text{tree}}, P_{\text{user}}, P_{\text{prompt(E)}}, C).
\end{align}}%

\noindent The training is conducted in two stages. In Stage 1, $\Theta_{\text{user}}$ and $\Theta_{\text{tree}}$ are pretrained via self-supervised response generation. In Stage 2, we initialize $\Theta_{\text{prompt}}$ randomly and jointly optimize $\Theta_{\text{user}}$, $\Theta_{\text{tree}}$, and $\Theta_{\text{prompt}}$. For the recommendation task, we compute:
{\small\begin{align}
    \mathbf{O}     & = \operatorname{Pooling}(f(\widetilde{C}_{\text{rec}} \mid \Theta)), \quad \hat{R} = \operatorname{softmax}(\mathbf{O} \mathbf{I}^{\top}), \\
    L_{\text{rec}} & = -\sum_{j=1}^{N} \sum_{i=1}^{M} [Y_j^i \log \hat{R}_j^i + (1 - Y_j^i) \log (1 - \hat{R}_j^i)],                                            \\
    L_{\text{all}} & = L_{\text{rec}} + \alpha L_{\text{user}} + \beta L_{\text{align}},
\end{align}}%
\noindent where $f(\cdot \mid \Theta)$ is the output of DialoGPT with input tokens and parameters. Pooling can be mean, max, or [CLS] embedding. $\alpha$ and $\beta$ are weighting hyperparameters. For the conversation task, we follow a similar two-stage training strategy, replacing  $L_{\text{rec}}$ with standard generation loss $L_{\text{conv}}$.

\section{Experiments}
\begin{table*}[ht]
    \centering
    \small
    \setlength{\tabcolsep}{1.7mm}
    \begin{tabular}{lrrrrrr|rrrrrr}
        \toprule
                                  & \multicolumn{6}{c|}{\textbf{INSPIRED}} & \multicolumn{6}{c}{\textbf{ReDial}}                                                                                                                                                                             \\
        \textbf{Model}            & \textbf{R@10}                          & \textbf{R@50}                       & \textbf{N@10}    & \textbf{N@50}    & \textbf{M@10}    & \textbf{M@50}    & \textbf{R@10} & \textbf{R@50} & \textbf{N@10} & \textbf{N@50} & \textbf{M@10} & \textbf{M@50} \\
        \midrule
        ReDial                    & 0.106                                  & 0.223                               & 0.049            & 0.075            & 0.031            & 0.037
                                  & 0.050                                  & 0.186                               & 0.024            & 0.053            & 0.015            & 0.021                                                                                                            \\
        KBRD                      & 0.151                                  & 0.278                               & 0.102            & 0.128            & 0.086            & 0.091
                                  & 0.189                                  & 0.372                               & 0.101            & 0.141            & 0.074            & 0.082                                                                                                            \\
        KGSF                      & 0.178                                  & 0.294                               & 0.109            & 0.133            & 0.088            & 0.093
                                  & 0.177                                  & 0.369                               & 0.094            & 0.137            & 0.069            & 0.078                                                                                                            \\
        TG-ReDial                 & 0.173                                  & 0.331                               & 0.110            & 0.144            & 0.091            & 0.098
                                  & 0.179                                  & 0.353                               & 0.101            & 0.140            & 0.078            & 0.086                                                                                                            \\
        UNICRS                    & 0.262                                  & 0.406                               & 0.159            & 0.193            & 0.131            & 0.138
                                  & 0.213                                  & 0.414                               & 0.119            & 0.163            & 0.090            & 0.100                                                                                                            \\
        KERL                      & 0.206                                  & 0.380                               & 0.128            & 0.168            & 0.113            & 0.117
                                  & 0.216                                  & 0.421                               & 0.122            & 0.161            & 0.087            & 0.097                                                                                                            \\
        DCRS                      & 0.267                                  & 0.410                               & 0.162            & 0.196            & 0.132            & 0.139
                                  & 0.217                                  & 0.426                               & 0.119            & 0.166            & 0.090            & 0.101                                                                                                            \\
        \midrule
        \textbf{PCRS-TKA}         & \textbf{0.278}*                        & \textbf{0.438}*                     & \textbf{0.187}*  & \textbf{0.222}*  & \textbf{0.158}*  & \textbf{0.166}*
                                  & \textbf{0.221}*                        & \textbf{0.432}*                     & \textbf{0.123}*  & \textbf{0.170}*  & \textbf{0.093}*  & \textbf{0.103}*                                                                                                  \\
        \textit{Improvement (\%)} & \textit{4.12\%}                        & \textit{6.83\%}                     & \textit{15.43\%} & \textit{13.27\%} & \textit{19.70\%} & \textit{19.42\%}
                                  & \textit{1.84\%}                        & \textit{1.41\%}                     & \textit{3.36\%}  & \textit{2.41\%}  & \textit{3.33\%}  & \textit{1.98\%}                                                                                                  \\
        \bottomrule
    \end{tabular}
    \caption{Evaluation results for recommendation task across two datasets (INSPIRED and ReDial). ``R'', ``N'', and ``M'' refer to ``recall'', ``ndcg'', and ``mrr'', respectively.}
    \label{tab:rec}
\end{table*}

\subsection{Experimental Setup}
\textbf{Dataset.} We conducted experiments on the ReDial~\cite{liDeepConversationalRecommendations2019} and INSPIRED~\cite{hayatiINSPIREDSociableRecommendation2020} datasets. Both datasets were constructed using the Amazon Mechanical Turk (AMT) platform. The ReDial dataset contains 10,006 dialogues, with an average of 18.2 utterances per dialogue and 14.5 words per utterance, covering 6,281 movies. In contrast, the INSPIRED dataset includes 1,001 dialogues, with an average of 35.7 utterances per dialogue and 19 words per utterance, encompassing 1,472 movies. In the experiment, we split each dataset into training, validation, and test sets in an 8:1:1 ratio. For each dialogue, we incrementally added one round of utterances starting from the first round to create new data, thereby expanding the dataset.

\noindent\textbf{Knowledge Graph.} DBpedia~\cite{auerDBpediaNucleusWeb2007} is a large-scale KG extracted from the structured content of Wikipedia. It contains 5,040,986 high-frequency entities with their corresponding 927 relations and 24,267,796 triplets. Since the entire DBpedia graph is too large, we collected all entities in the dataset corpus using TagMe tool and extracted their one-hop triples, forming the subgraph used as the external KG in our experiment.

\noindent \textbf{Evaluation Metrics.} We conducted two types of evaluations: recommendation and conversation. For the recommendation task, we used recall@k (k=10, 50), ndcg@k (k=10, 50), and mrr@k (k=10, 50) as metrics. For the conversation task, we employed both automatic and manual evaluations. The automatic evaluation used word-level distinct-n (n=2, 3, 4) to measure response diversity. Additionally, for manual evaluation, we randomly selected 100 conversations and their model-generated responses and invited ten annotators to score them. The evaluation assessed four aspects: \textit{Fluency}, \textit{Informativeness(Inform.)}, \textit{Consistency}, and \textit{Accuracy}, with scores ranging from 0 to 5. Further details are provided in our public code repository.

\noindent \textbf{Benchmark Models.} We compared PCRS-TKA with several state-of-the-art models, including:
\textbf{ReDial}: Integrates an autoencoder-based recommendation module with an HRED-based conversation module.
\textbf{KBRD}: Utilizes DBpedia to enhance entity representations, combining a self-attention recommendation module with a Transformer-based conversation module.
\textbf{KGSF}: Integrates ConceptNet~\cite{speerConceptNet55Open2018} and DBpedia, using mutual information maximization to align their semantic spaces.
\textbf{TG-ReDial}~\cite{zhouTopicGuidedConversationalRecommender2020}: Introduces a topic prediction task, using SASRec~\cite{kangSelfAttentiveSequentialRecommendation2018} for recommendation, BERT~\cite{devlinBERTPretrainingDeep2019} for topic prediction, and GPT-2~\cite{gaoMakingPretrainedLanguage2021} for response generation.
\textbf{UniCRS}: Uses prompt learning to guide a PLM for both tasks, integrating DBpedia information into the prompt.
\textbf{KERL}: Introduces a Wikipedia KG with entity descriptions and a PLM encoder for better entity representations.
\textbf{DCRS}: Incorporates a knowledge-aware retriever to collect selective analogues from dialogue histories.

\noindent \textbf{Implementation Details.} We conducted the experiment on one single NVIDIA-V100 GPU. We used grid search to choose the hyperparameters. After searching, we used AdamW with epsilon set to 0.01, learning rate set to 5e-4 for first-stage pre-training, and 1e-4 for second-stage training for both recommendation and conversation tasks. The batch size was set to 64 for the recommendation task and 8 for the conversation task. The soft prompt token length was set to 10 for the recommendation task and 20 for the conversation task. For all baseline methods, hyperparameters were also tuned using grid search. More details can be found in our public code repository.

\subsection{Overall Performance}
\textbf{Recommendation Task.} As shown in Table \ref{tab:rec}, our model consistently outperforms all baseline models on both INSPIRED and ReDial datasets. Specifically, it achieves significant improvements over the best-performing baseline, DCRS, with improvements of up to 19.70\% on INSPIRED and 3.33\% on ReDial in key metrics like mrr@10. The larger improvements on the INSPIRED dataset are likely due to the greater number of utterances per dialogue, which enables PCRS-TKA to capture more complex user preferences. Those using external KGs, such as KBRD and KGSF, generally outperform the basic ReDial model. And KGSF outperforms KBRD for the integration of ConceptNet. However, these methods show limited improvement compared to UniCRS, which leverages PLMs and prompt learning. KERL introduces a Wikipedia KG with entity descriptions and DCRS uses a knowledge-aware retriever. Our model, utilizes PLMs to process tree-structured KGs and incorporates user preference extraction from multi-turn dialogues, leading to its superior performance over all baselines.

\begin{table}
    % \color{red}
    \centering
    \small
    \setlength{\tabcolsep}{0.23mm}
    \begin{tabular}{lrrrrrr}
        \toprule
                                  & \multicolumn{3}{c}{\textbf{INSPIRED}} & \multicolumn{3}{c}{\textbf{ReDial}}                                                                             \\
        \textbf{Model}            & \textbf{Dist-2}                       & \textbf{Dist-3}                     & \textbf{Dist-4}  & \textbf{Dist-2}  & \textbf{Dist-3}  & \textbf{Dist-4}  \\
        \midrule
        ReDial                    & 0.313                                 & 1.237                               & 2.562            & 0.070            & 0.279            & 0.643            \\
        KBRD                      & 0.567                                 & 2.017                               & 3.621            & 0.094            & 0.488            & 1.004            \\
        KGSF                      & 0.657                                 & 2.822                               & 5.992            & 0.110            & 0.656            & 1.729            \\
        TG-ReDial                 & 0.778                                 & 2.825                               & 5.511            & 1.016            & 1.487            & 1.642            \\
        UniCRS                    & 3.949                                 & 6.004                               & 7.082            & 0.899            & 1.267            & 1.390            \\
        KERL                      & 3.102                                 & 5.194                               & 6.548            & 0.797            & 1.511            & 1.783            \\
        DCRS                      & 4.383                                 & 6.140                               & 7.312            & 0.922            & 1.313            & 1.422            \\
        \midrule
        \textbf{PCRS-TKA}         & \textbf{6.805}*                       & \textbf{9.906}*                     & \textbf{10.804}* & \textbf{1.162}*  & \textbf{1.676}*  & \textbf{1.845}*  \\
        \textit{Improvement (\%)} & \textit{55.23\%}                      & \textit{61.33\%}                    & \textit{47.74\%} & \textit{26.02\%} & \textit{27.65\%} & \textit{29.74\%} \\
        \bottomrule
    \end{tabular}
    \caption{Evaluation results for conversation task across two datasets (INSPIRED and ReDial). ``Dist'' refers to ``distinct''.}
    \label{tab:conv}
\end{table}

\noindent \textbf{Conversation Task.} Table~\ref{tab:conv} presents the evaluation results for the conversation task, where PCRS-TKA outperforms all baselines across both datasets in distinct-n metrics. Specifically, it improves up to 55.23\% in distinct-2 on the INSPIRED dataset and 29.74\% in distinct-4 on the ReDial dataset compared to the best baseline, DCRS. These results demonstrate that PCRS-TKA generates more diverse responses, crucial for engaging and natural conversations. While distinct-n metrics measure lexical diversity, they don’t fully capture dialogue quality. Table~\ref{tab:conv2} presents human evaluation results, where PCRS-TKA outperforms all baselines in fluency, informativeness, consistency, and accuracy. Our model achieves improvements of 6.36\% in fluency, 14.53\% in informativeness, 3.01\% in consistency, and 1.24\% in accuracy over DCRS. The superior performance is driven by two factors: integrating structured dialogue-specific knowledge from the KG and leveraging user preferences across multi-turn conversations for better alignment. While DCRS excels at retrieving contextualized information, PCRS-TKA captures conversational dynamics more effectively. UniCRS surpasses KGSF and ReDial by leveraging PLMs for text generation. KGSF benefits from external KGs but doesn’t fully capture the nuances of dialogue like PCRS-TKA does.

\begin{table}
    \centering
    \small
    \setlength{\tabcolsep}{1.21mm}
    \begin{tabular}{lrrrr}
        \toprule
        \textbf{Models}           & \textbf{Fluency} & \textbf{Inform.} & \textbf{Consistency} & \textbf{Accuracy} \\
        \midrule
        ReDial                    & 2.57             & 2.11             & 1.96                 & 3.04              \\
        KGSF                      & 3.08             & 1.98             & 1.86                 & 2.76              \\
        UniCRS                    & 3.43             & 3.36             & 2.98                 & 3.14              \\
        KERL                      & 3.47             & 3.44             & 3.28                 & 3.23              \\
        DCRS                      & 3.93             & 3.36             & 3.66                 & 3.07              \\
        \midrule
        \textbf{PCRS-TKA}         & \textbf{4.18}*   & \textbf{3.94}*   & \textbf{3.77}*       & \textbf{3.27}*    \\
        \textit{Improvement (\%)} & \textit{6.36\%}  & \textit{14.53\%} & \textit{3.01\%}      & \textit{1.24\%}   \\
        \bottomrule
    \end{tabular}
    \caption{Human evaluation results on INSPIRED dataset for conversation task.}
    \label{tab:conv2}
\end{table}

\begin{figure}[t]
    \centering
    \includegraphics[width=\linewidth]{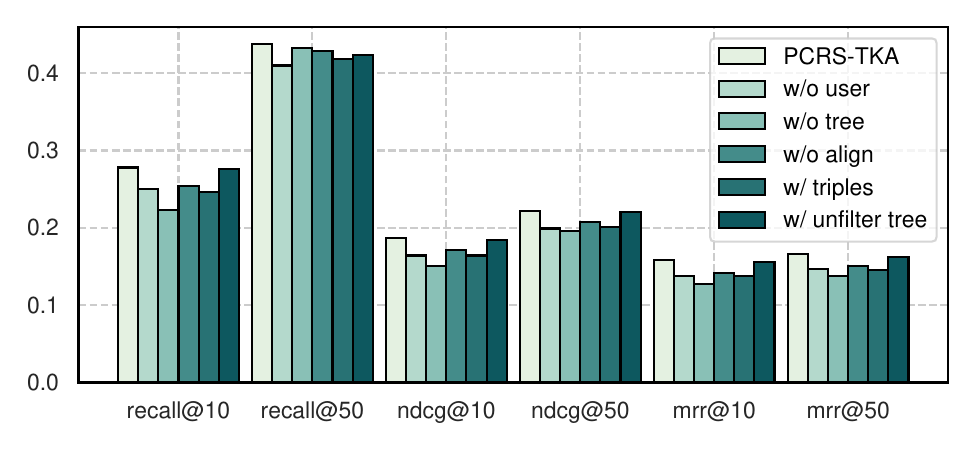}
    \caption{Ablation study on INSPIRED dataset for recommendation task.}
    \label{fig:ablation-inspired}
\end{figure}

\begin{table}[t]
    \centering
    \small
    \setlength{\tabcolsep}{0.4mm}
    \begin{tabular}{lrrrrrr}
        \toprule
        \textbf{Model}    & \textbf{R@10} & \textbf{R@50} & \textbf{N@10} & \textbf{N@50} & \textbf{M@10} & \textbf{M@50} \\
        \midrule
        PCRS-TKA (BERT)   & 0.254         & 0.441         & 0.174         & 0.214         & 0.148         & 0.156         \\
        -all (BERT)       & 0.250         & 0.426         & 0.162         & 0.201         & 0.138         & 0.143         \\
        \midrule
        PCRS-TKA (GPT2)   & 0.258         & 0.430         & 0.162         & 0.204         & 0.140         & 0.147         \\
        -all (GPT2)       & 0.230         & 0.422         & 0.153         & 0.196         & 0.129         & 0.138         \\
        \midrule
        PCRS-TKA (origin) & 0.278         & 0.438         & 0.187         & 0.222         & 0.158         & 0.166         \\
        -all (origin)     & 0.262         & 0.406         & 0.159         & 0.193         & 0.131         & 0.138         \\
        \bottomrule
    \end{tabular}
    \caption{Generalizability analysis on INSPIRED for recommendation task.}
    \label{tab:ablation_plms}
\end{table}

\begin{figure*}[ht]
    \centering
    \includegraphics[width=\textwidth]{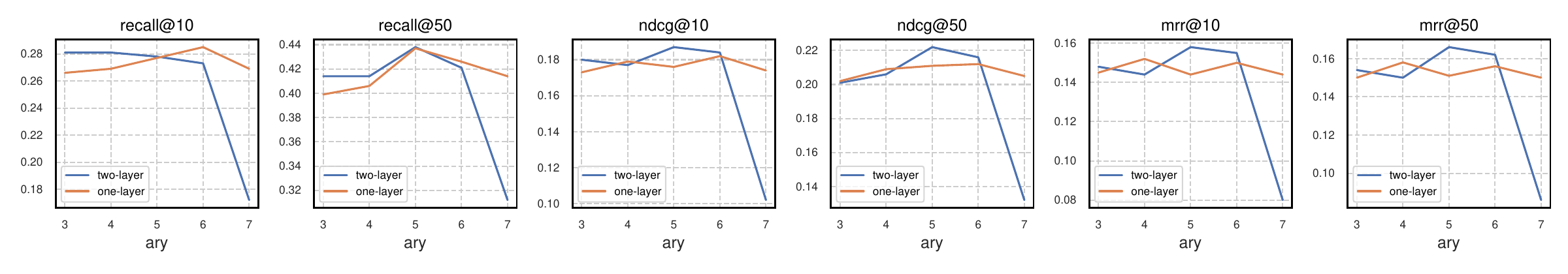}
    \caption{Model performance comparison with varing degree and depth of knowledge tree on INSPIRED dataset. }
    \label{fig:tree}
    \includegraphics[width=\textwidth]{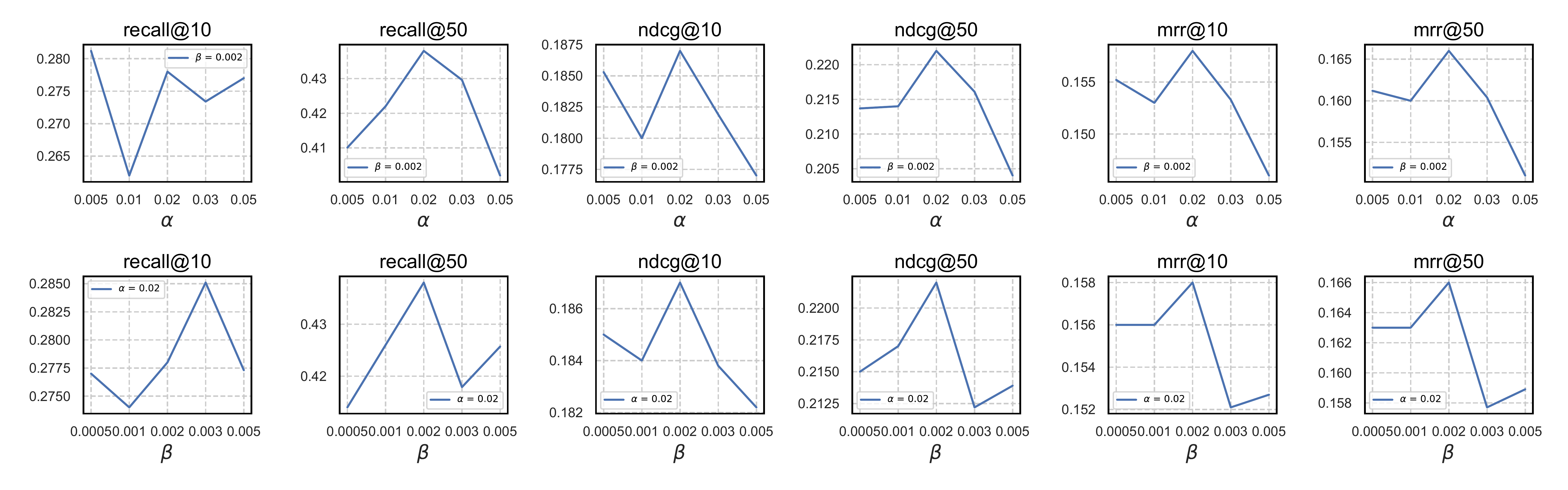}
    \caption{Model performance comparison with varing alpha and beta on INSPIRED dataset. }
    \label{fig:loss}
\end{figure*}

\subsection{Ablation Study \label{ablation}}
We performed ablation experiments to evaluate the contribution of each component in our approach. Specifically, we excluded the knowledge tree enhanced prompts, user preference prompts, and the multimodal information integration module, separately, during both the pre-training and training phases. As shown in Figure \ref{fig:ablation-inspired}, removing any component leads to performance degradation, confirming that all components are crucial for improving the recommendation task. To further validate the design of our core knowledge tree module, we conducted deeper ablations. For instance, replacing the hierarchical tree with simple unordered triples caused a significant performance drop, confirming that the tree structure is critical for preserving a clear reasoning path for the PLM, a finding that aligns with prior work like TREA~\cite{li-etal-2023-trea}. Similarly, removing our context-aware filtering mechanism also degraded performance. This result, combined with our hyperparameter analysis showing that performance declines as tree degree increases, provides strong evidence that our RAG-style filtering is significant for reducing noise. The ability to isolate such a relevant, filtered subgraph also gives our framework strong potential for robustness against dynamic or noisy KGs.

\subsection{Generalizability Analysis}
To evaluate the generalizability of PCRS-TKA across various PLMs, we conducted experiments by substituting RoBERTa with BERT and DialoGPT with GPT-2 to assess whether the framework could be seamlessly adapted to different model architectures while maintaining its effectiveness. The experimental results, presented in Table \ref{tab:ablation_plms}, demonstrate the consistent performance of the proposed modules in PCRS-TKA across these alternative PLMs. These findings underscore the flexibility and adaptability of our approach, confirming its applicability to diverse model architectures.

Notably, in principle, we could use larger LLMs to replace RoBERTa and DialoGPT in PCRS-TKA. However, LLMs require substantial computational resources, and prior studies~\citep{schick2021itsjustsizematters, subramanian2025smalllanguagemodelsslms} have discussed that in tasks with modest semantic complexity and generalization demands, such as movie recommendations, LLMs may not surpass PLMs. In contrast, PLMs like RoBERTa and DialoGPT strike a balance between performance and efficiency, making them more practical in resource-constrained environments. In this paper, we focus on leveraging PLMs to achieve high-quality recommendations while maintaining efficiency.

\subsection{Hyperparameter Sensitivity Analysis \label{hyper}}
\noindent \textbf{Analysis on the degree and depth of the knowledge tree.}
We conduct parameter-tuning experiments to study the impact of the knowledge tree’s depth and degree on recommendation performance. As results shown in Figure~\ref{fig:tree}, one-layer and two-layer trees yield similar optimal recommendation results. This indicates that a two-hop path captures sufficient information—e.g., (movie, starring, actor). For one-layer trees, the model’s performance remains stable as the tree degree changes. However, for two-layer trees, performance declines sharply as the degree increases.  We hypothesize this is because deeper and wider trees are more likely to introduce contextually irrelevant entities, which act as noise. This noise can dilute the signal from the more critical first-hop information and cause the model to over-focus on spurious or less relevant multi-hop relationships. This finding underscores the importance of our context-aware filtering and suggests that the tree's depth and degree should be carefully adapted to the KG structure and dialogue context to minimize noise.

\noindent \textbf{Analysis on loss balancing.} The recommendation task loss function includes two hyperparameters, $\alpha$ and $\beta$, to balance its components. The parameter $\alpha$ controls collaborative signal weight: as $\alpha$ increases, user preference influence grows, enhancing performance until excessive emphasis over shadows dialogue context, leading to a decline. Similarly, $\beta$ tunes alignment: as $\beta$ rises, alignment improves performance up to a point, beyond which can disrupt information learned by entity embeddings via RGCN, while a smaller proportion facilitates better integration of entity and tree embeddings. As shown in Figure~\ref{fig:loss}, optimal performance occurs with $\alpha \approx 0.02$ and $\beta \approx 0.002$. These values remained stable across datasets, indicating consistent performance trends and an efficient tuning process.

\section{Conclusion}
In this paper, we presented PCRS-TKA, a novel framework for conversational recommendation featuring a synergistic architecture that integrates pretrained language models with knowledge graphs. Our approach uniquely augments static, global KG embeddings with dynamic, context-aware knowledge trees constructed in a RAG-style, enabling direct and fine-grained reasoning by the PLM. Furthermore, it explicitly models collaborative preferences by learning to encode this signal into the prompts via an auxiliary task. Extensive experiments demonstrate that this design allows PCRS-TKA to consistently outperform strong baselines in both recommendation quality and dialogue generation.

\section{Acknowledgments}
This work was supported in part by the Natural Science Foundation of Anhui Province (Grant No. 2508085QF211), the National Key R\&D Program of China (Grant No. 2023YFF0725001), the National Natural Science Foundation of China (Grant Nos. 62506348, 92370204, 62406141), the Guangdong Basic and Applied Basic Research Foundation (Grant No. 2023B1515120057), the Key-Area Special Project of Guangdong Provincial Ordinary Universities (Grant No. 2024ZDZX1007), the CCF-1688 Yuanbao Cooperation Fund (Grant No. CCF-Alibaba2025005), the China Postdoctoral Science Foundation (Grant No. GZC20252740), and the Education Bureau of Guangzhou.

\bibliography{aaai2026}

\end{document}